\documentclass[10pt,twocolumn,letterpaper]{article}

\usepackage[pagenumbers]{cvpr} 

\definecolor{cvprblue}{rgb}{0.21,0.49,0.74}
\usepackage[pagebackref,breaklinks,colorlinks,allcolors=cvprblue]{hyperref}
\usepackage{multirow}
\usepackage{multicol}
\usepackage{enumitem}
\usepackage{graphicx} 
\usepackage{booktabs}
\usepackage{url}

\def\paperID{17876} 
\def\confName{CVPR}
\def\confYear{2026}

\title{UniSER: A Foundation Model for Unified Soft Effects Removal}

\author{
    Jingdong Zhang$^{1,2}$ \quad
    Lingzhi Zhang$^2$ \quad
    Qing Liu$^2$ \quad
    Mang Tik Chiu$^2$ \quad
    Connelly Barnes$^2$ \\
    Yizhou Wang$^2$ \quad
    Haoran You$^2$ \quad
    Xiaoyang Liu$^2$ \quad
    Yuqian Zhou$^2$ \quad
    Zhe Lin$^2$ \\
    Eli Shechtman$^2$ \quad
    Sohrab Amirghodsi$^2$ \quad
    Xin Li$^1$ \quad
    Wenping Wang$^1$ \quad
    Xiaohang Zhan$^2$ \\
    $^1$Texas A\&M University \quad
    $^2$Adobe Research \\
    {
    \tt\small \{jdzhang, xinli, wenping\}@tamu.edu,
    \{lingzzha, xzhan\}@adobe.com
    }
}

\begin{document}

\twocolumn[{
    \renewcommand\twocolumn[1][]{#1}
    \maketitle
    
    \begin{center}
        \centering
        \includegraphics[width=0.98\textwidth]{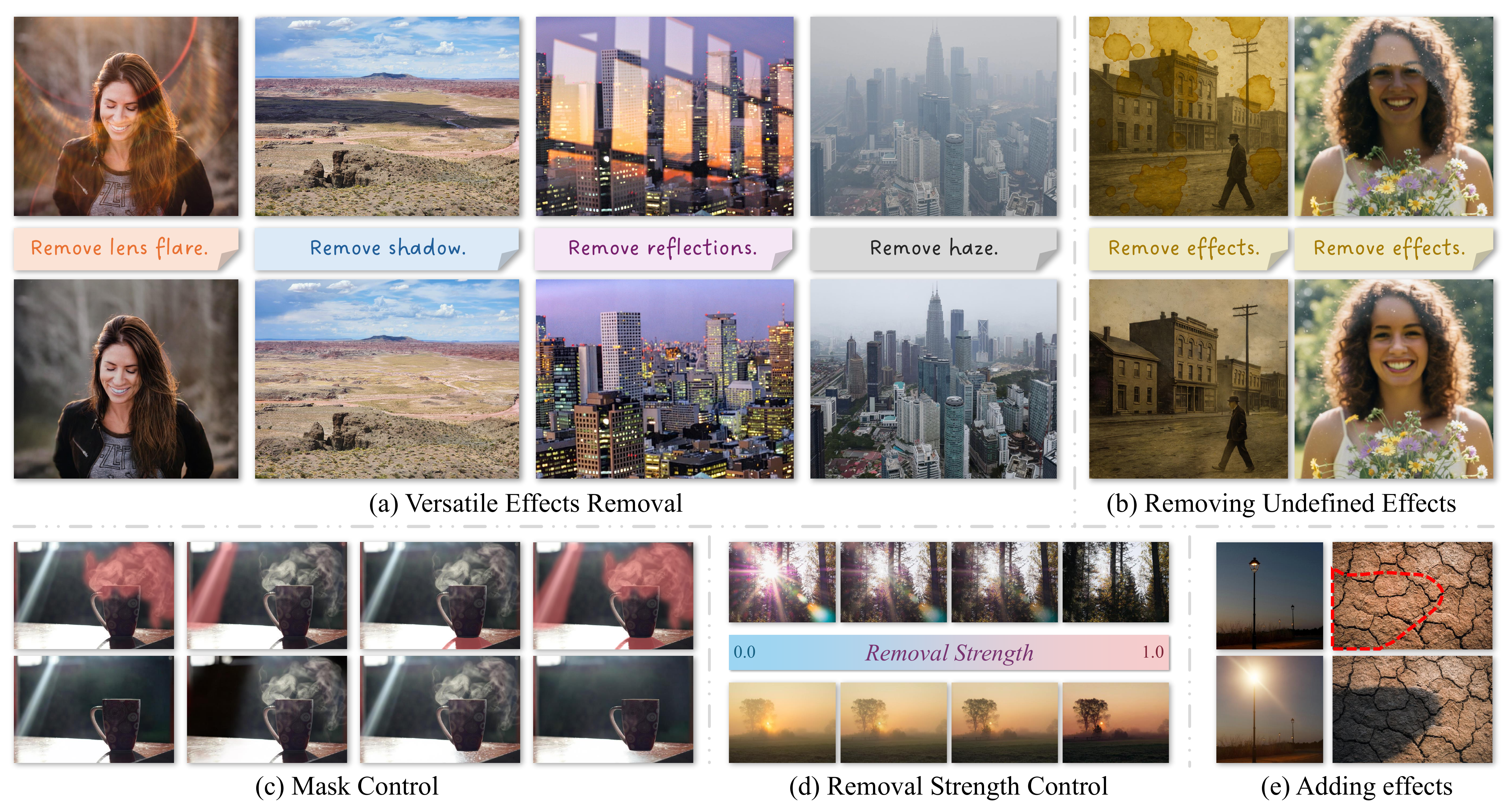}
        \vspace{-1.8mm}
        \captionof{figure}{
        Our UniSER eliminates multiple challenging (a) and even undefined (b) soft effects from in-the-wild images while preserving background identities.
        Besides, UniSER supports precise pixel mask control (c), and removal strength control (d), allowing for intuitive and fine-grained restoration tailored to specific user needs. The framework is also capable of adding effects in the given region (e). Masks are global by default if not shown. \textbf{A demo video is included in the supplementary materials.}
        }
        \label{fig:teaser}
    \end{center}
}]

\begin{abstract}
Digital images are often degraded by soft effects such as lens flare, haze, shadows, and reflections, which reduce aesthetics even though the underlying pixels remain partially visible. 
The prevailing works address these degradations in isolation, developing highly specialized, specialist models that lack scalability and fail to exploit the shared underlying essences of these restoration problems. While specialist models are limited, recent large-scale pretrained generalist models offer powerful, text-driven image editing capabilities. while recent general-purpose systems (\textit{e.g.}, GPT-4o, Flux Kontext, Nano Banana) require detailed prompts and often fail to achieve robust removal on these fine-grained tasks or preserve identity of the scene. 
Leveraging the common essence of soft effects, \textit{i.e.}, semi-transparent occlusions, we introduce a foundational versatile model UniSER, capable of addressing diverse degradations caused by soft effects within a single framework.
Our methodology centers on curating a massive 3.8M-pair dataset to ensure robustness and generalization, which includes novel, physically-plausible data to fill critical gaps in public benchmarks, and a tailored training pipeline that fine-tunes a Diffusion Transformer to learn robust restoration priors from this diverse data, integrating fine-grained mask and strength controls. This synergistic approach allows UniSER to significantly outperform both specialist and generalist models, achieving robust, high-fidelity restoration in the wild.

\end{abstract}    
\section{Introduction}
\label{sec:intro}

Images captured in real-world environments inevitably suffer from degradations. A common class of such “soft” effects includes optical phenomena (e.g., lens flare, reflections) and atmospheric conditions (e.g., haze, fog). These effects corrupt scene radiance additively or multiplicatively, degrading contrast, color fidelity, and fine details~\citep{le2019shadow, wan2017benchmarking}. Consequently, image quality and visibility are compromised, and in severe cases, occlusions cause irreversible information loss, rendering recovery fundamentally ill-posed~\citep{he2010single, wu2021train, le2019shadow}.

To restore image structures, most existing works address each degradation type separately. For instance, dehazing has progressed from prior-based methods such as the Dark Channel Prior (DCP)~\citep{he2010single} to deep networks estimating scattering parameters or directly predicting clean images~\citep{li2017aod,song2023vision,chen2021psd,engin2018cycle,chen2019gated}. Similarly, shadow, flare, and reflection removal adopt task-specific designs~\citep{le2019shadow,dong2024shadowrefiner,wu2021train,xue2025dfdnet,zhu2024revisiting,wan2017benchmarking}, relying on physical modeling, layer decomposition, or elaborate data and network strategies to mitigate ill-posedness. While such methods achieve strong task-specific performance, recent works~\citep{chen2025unirestore,li2020all,potlapalli2023promptir} attempt to unify multiple degradations within one framework. Yet these models remain limited in scalability and robustness when facing extreme, diverse real-world conditions. This motivates the development of foundation models trained on large-scale data to achieve stronger generalization and resilience in the wild.

Concurrently, the rise of powerful foundation models like GPT-4o~\citep{hurst2024gpt} and Nano Banana (Gemini 2.5 Flash Image)~\citep{comanici2025gemini,google:gemini2.5flash_image} has introduced general-purpose, text-driven image generation/editing based on Multi-modal Large Language Models (MLLMs). These models can interpret complex prompts and perform realistic edits. However, for fine-grained tasks like soft effect removal, they exhibit significant limitations. Their performance is often unstable and heavily reliant on meticulously crafted text prompts. More critically, they lack the precise, pixel-wise control required for high-fidelity restoration and identity preservation. Treating soft effect removal as a general inpainting task leads to the alteration of local image structures or the identity of objects in the scene, which makes them unreliable for professional photo editing and critical computer vision pipelines.

Despite their diverse appearances, effects such as lens flare, haze, reflections, and shadows share the same intrinsic property: they are semi-transparent occlusions that degrade the image, but do not fully destroy the underlying scene identity. This shared property unifies them as a challenging decomposition problem (e.g.~\cite{ghodesawar2023deflare,zhou2023improving,zhou2025image,zhou2024difflare} use dehaze models as strong comparative baselines for lens flares removal). To this end, we define a unified and extensible task, termed Soft Effects Removal (SER) to invert all these diverse degradation processes. This task is highly challenging. First, these effects are typically entangled with the scene itself, rather than merely superimposed as simple overlays. Second, the local image structures, and even pixel-level identities, should be precisely preserved. Third, regions that are fully occluded or invisible (\textit{e.g.}, overexposed areas in lens flare or areas covered by extremely dense haze) must be plausibly reconstructed. 

To effectively tackle these challenges, we introduce \textbf{UniSER} (Fig.~\ref{fig:teaser} (a) \& (b)), a data-centric versatile model for Soft Effects Removal. Our method is built upon two key points. First, we curated a large-scale dataset of approximately 3.8M balanced, high-quality, pixel-aligned image pairs. By unifying existing open-source datasets and augmenting them with extra real-world and synthetic data, we provide the precise supervision our model needs to learn content invariance. Second, as shown in Fig.~\ref{fig:teaser} (c) \& (d), we implemented fine-grained user controls, including pixel-level masks to define the removal area and strength levels to modulate the removal strength, making the process highly controllable. Beyond restoration, UniSER can also perform aesthetic edits, such as enhancing existing effects or generating new, realistic ones on clean images (Fig.~\ref{fig:teaser} (e)). Our method achieves state-of-the-art results on multiple public benchmarks and demonstrates significantly better generalization on in-the-wild testing data.


In summary, our main contributions can be summarized as follows:
\begin{itemize}[nosep,leftmargin=*]
    \item \textbf{A Large-Scale Dataset for Generalization:} We curated a large-scale dataset of$\sim$3.8M image pairs, providing vast data distribution for strong generalization on challenging in-the-wild data.
    \item \textbf{A Versatile SER Model:} Trained on the curated dataset, a foundational versatile model UniSER achieves removing multiple challenging soft effects in the wild with state-of-the-art performance and surpasses much larger general-purpose models such as Nano Banana.
    \item \textbf{Controllable Editing:} Developed fine-grained user controls for SER tasks, including spatial masks and strength levels, to enable precise and controllable effect removal.
\end{itemize}
\section{Related Work}
\label{sec:related_work}

\subsection{Isolated Effects Removal}
\noindent\textbf{Lens flare removal.}
Previous learning-based methods improved data synthesis by considering camera ISP to enhance realism and generalization~\citep{zhou2023improving, zhou2025image}. Concurrently, architectural innovations emerged, including self-supervised methods to disentangle co-occurring flares~\citep{he2025disentangle}, while others explicitly separated light source preservation from flare removal using dedicated detection modules~\citep{ghodesawar2023deflare}, and networks leveraging both spatial and frequency domains~\citep{vasluianu2024sfnet}. More recently, large pretrained Latent Diffusion Models (LDMs) are adpated to leverage their powerful generative priors~\citep{zhou2024difflare}. The development of these methods has also been heavily reliant on specialized datasets, from semi-synthetic ones~\citep{wu2021train}, Flare7K~\citep{dai2022flare7k}, to real-world paired datasets~\citep{dai2024mipi}.

\noindent\textbf{Reflection removal.}
Early methods for single-image reflection removal (SIRR) focused on iterative refinement using edge maps~\citep{fan2017generic} or recurrent networks~\citep{yang2018seeing, li2020single}. Subsequent research shifted towards improving training data realism by learning non-linear blending~\citep{wen2019single}, employing physically-based rendering~\citep{kim2020single}, and modeling glass absorption~\citep{zheng2021single}. Architectural innovations followed, introducing location-aware modules~\citep{dong2021location} and advanced attention mechanisms~\citep{huang2025single, zhang2025pa} to better distinguish between layers. More recent paradigms reduce reliance on paired data through unsupervised deep image priors~\citep{rahmanikhezri2022unsupervised}, RAW data simulation~\cite{kee2025removing}, or by using Diffusion Models to generate guiding prompts~\citep{wang2024promptrr}. This progress has been underpinned by the creation of key real-world benchmarks like $SIR^2$~\citep{wan2017benchmarking} and the large-scale RRW dataset~\citep{zhu2024revisiting}.

\noindent\textbf{Shadow removal.}
Initial approaches to shadow removal relied on traditional physical priors and optimization frameworks~\citep{guo2011single, zhang2015shadow}. The advent of deep learning introduced end-to-end models like DeshadowNet~\citep{qu2017deshadownet} and methods that decomposed images into shadow-free and matte layers~\citep{le2019shadow}. Subsequent architectural advancements included using Generative Adversarial Networks (GANs) for joint detection and removal~\citep{wang2018stacked}, fusing synthetic exposure pairs~\citep{fu2021auto}, and learning via shadow generation~\citep{liu2021shadow}. More recent trends focus on eliminating the dependency on explicit shadow masks, utilizing mask-free transformers~\citep{dong2024shadowrefiner} or reformulating the problem as a dense prediction task~\citep{lin2025densesr}. The progress in this field has been propelled by benchmarks like SRD~\citep{qu2017deshadownet}, ISTD~\citep{wang2018stacked}, and the newer high-resolution WSRD dataset~\citep{vasluianu2023wsrd}.

\noindent\textbf{Haze removal.}
Single-image dehazing evolved from early methods based on statistical priors like the Dark Channel Prior (DCP)~\citep{he2010single} to data-driven deep learning. Initial deep learning works included lightweight end-to-end networks~\citep{li2017aod}, hybrid models that learned priors for traditional optimization~\citep{yang2018proximal}, and unpaired training with GANs to address data scarcity~\citep{engin2018cycle}. Architectural innovations, such as gated context aggregation~\citep{chen2019gated} and Vision Transformers~\citep{song2023vision}, were later introduced to better handle non-uniform haze. Recent efforts focus on closing the synthetic-to-real domain gap by generating more physically plausible training data~\citep{chen2021psd} or leveraging diffusion models for realistic haze synthesis~\citep{wang2025learning}. This progress has been consistently driven by the development of comprehensive benchmarks~\citep{li2018benchmarking,zhang2024lmhaze,islam2024hazespace2m}.

Apart from them, some works delve into All-In-One (AIO) methods to restore image quality from multiple degradations within a multi-task model~\citep{li2020all,potlapalli2023promptir,chen2025unirestore,jiang2024autodir,rajagopalan2025awracle,tian2025degradation,liu2024diff,zheng2024selective,cuibio}. Despite the achievements from all these methods, key challenges persist including the limited diversity in datasets, while current methods still struggle with scalable training with robust generalization abilities, as well as handling more challenging types of challenging soft effects requiring semantic-awareness.

\subsection{Prompt-based Image Editing}
\vspace{-1mm}
Prompt-based image editing originated from diffusion models, enabled by deterministic inversion techniques like DDIM~\citep{song2020denoising} that map real images to an editable latent space. Initial methods controlled edits by manipulating internal model structures, such as altering cross-attention maps to preserve layout~\citep{hertz2022prompt} or fine-tuning the entire model on a single image for complex, non-rigid changes~\citep{kawar2023imagic}. The field has since evolved towards more direct user control, with models trained to follow natural language instructions~\citep{brooks2023instructpix2pix} or allow for interactive, point-based spatial adjustments~\citep{shi2024dragdiffusion}. This shift towards more precise, semantic editing is increasingly powered by the advanced contextual understanding of Multimodal Large Language Models (MLLMs)~\citep{hurst2024gpt,comanici2025gemini,bai2025qwen2}. However, current approaches still often lack fine-grained pixel control and can struggle to perfectly preserve the subject’s identity during transformation.
\begin{table*}[h]
\vspace{-4mm}
\caption{Summary of datasets curated for UniSER training. ``$\dagger$" represents the datasets curated by us, ``*'' represents the datasets which we re-synthesis effects with our own algorithm.}
\vspace{-2mm}
\label{tab:dataset_summary}
\centering
\setlength{\tabcolsep}{1.8mm}{\scalebox{0.75}{
\begin{tabular}{llllr}
\toprule
\textbf{Task} & \textbf{Dataset} & \textbf{Type} & \textbf{Description} & \textbf{Pairs} \\ \midrule
\multirow{2}{*}{Lens flare} & FlareReal600~\citep{dai2024mipi} & Real-World & Nighttime flares, Streetview, Cityscapes, Outdoor & 0.6k \\ 
 & HALO$\dagger$ & 3D Synthetic & Rendered, Various flares and scenes, Indoor \& Outdoor & 70k \\\midrule
\multirow{4}{*}{Shadow} & WSRD+~\citep{vasluianu2023wsrd} & Real-World & Object-level, Close-view, Rich texture, Complex shadows & 1k \\
 & ISTD+~\citep{wang2018stacked} & Real-World & Simple-shaped shadows, Monotonous scenes, Outdoor & 1.3k \\
 & SRD~\citep{qu2017deshadownet} & Real-World & Various scenes, Outdoor & 2.6k \\ 
 & LR-SRD$\dagger$ & Real-World & {Object-level, Close-view, Hard \& Soft shadow, Indoor \& Outdoor} & ~26k \\ \midrule
\multirow{7}{*}{Haze} & \parbox[t]{4cm}{Haze-R~\citep{ancuti2018ihaze,ancuti2018ohaze,ancuti2019dense,ancuti2020nh,ancuti2021ntire,ancuti2023ntire,ancuti2024ntire}} & Real-World & \parbox[t]{12cm}{Collection including: I-HAZE, O-HAZE, Dense-Haze, NH-Haze, etc., Homogeneous \& Non-Homogeneous, Indoor \& Outdoor} & 0.3k \\
 & REVIDE~\citep{zhang2021learning} & Real-World & Video Frames, Indoor & 1.9k \\ 
 & LM-Haze~\citep{zhang2024lmhaze} & Real-World & Multi-level haze, Homogeneous, Indoor & 5k \\
 & HAZESPACE*~\citep{islam2024hazespace2m} & 2D Synthetic & Multi-level haze, Vast range of scenes, Outdoor & 24$\times$70k \\
 & RESIDE*~\citep{li2018benchmarking} & 2D Synthetic & Multi-level haze, Indoor \& Outdoor & 290k \\
 & SYN-HAZE* & 2D Synthetic & {Multi-level haze, Synthetic scenes, Include extremely dense haze, Indoor \& Outdoor} & 24$\times$70k \\ \midrule
\multirow{4}{*}{Reflection} & RRW~\citep{zhu2024revisiting} & Real-World & Various scenes, Diverse glass and reflection types & 14.9k \\
 & POLAR-RR~\citep{lei2020polarized} &Real-World & Polarization-based, Indoor & 0.8k \\
 & RFC~\citep{lei2021robust} & Real-World & Flash-induced reflections & 5k \\ 
 & BDN~\citep{yang2018seeing} & 2D Synthetic & Linearly Blended, Public Image Sources & 50k \\\bottomrule
\end{tabular}
}}
\vspace{-2mm}
\end{table*}

\begin{figure*}[t]
  \centering
   \includegraphics[width=0.95\linewidth]{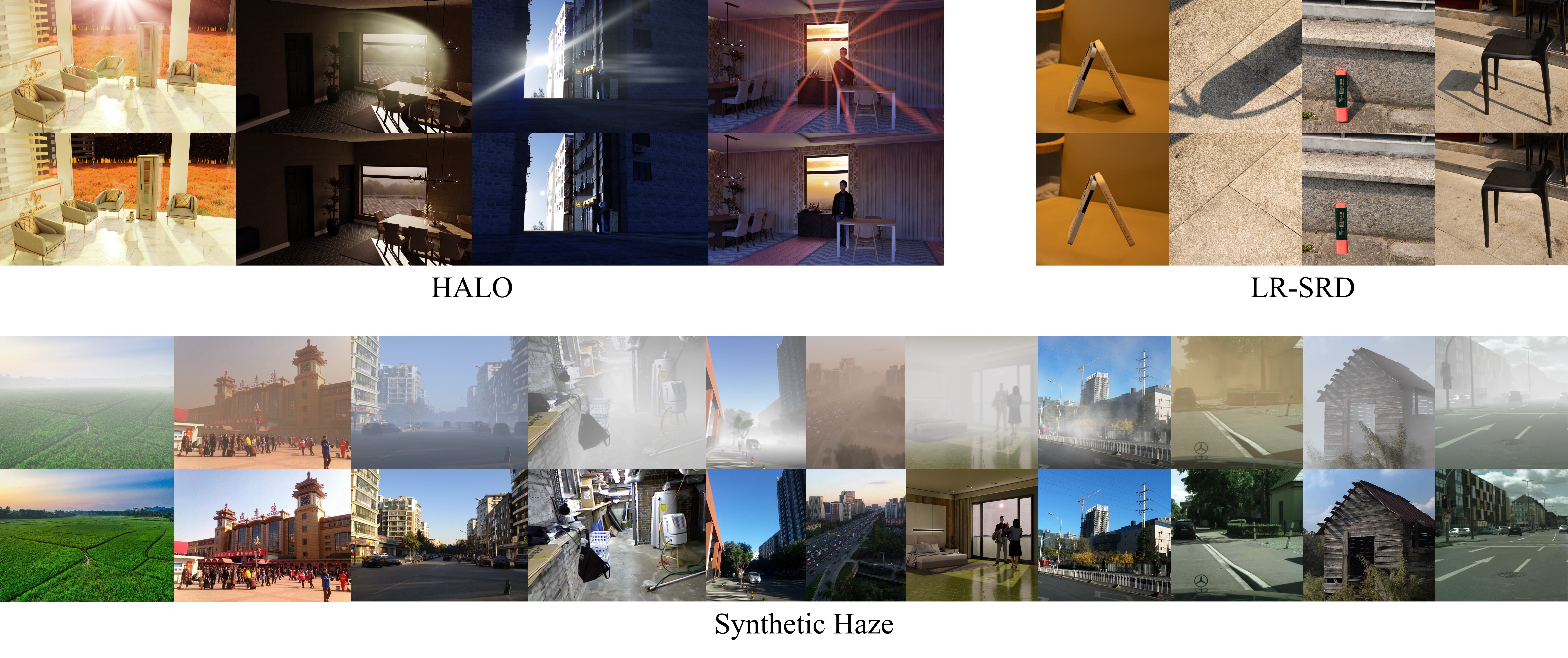}
   \vspace{-4mm}
   \caption{Visualization of our curated data samples and synthetic haze by our method.} 
   \vspace{-3mm}
   \label{fig:dataset_vis}
\end{figure*}

\vspace{-3mm}
\section{Methodology}
\vspace{-1mm}
\subsection{Data Curation}
\vspace{-1mm}
A powerful foundation model requires large-scale, high-quality, and diverse training data. To equip UniSER with robust generalization, we curated a comprehensive dataset by unifying pixel-aligned image pairs from four representative tasks: lens flare, shadow, haze, and reflection removal. This integration enables the model to learn a broad restoration representation while preserving content identity.

\noindent\textbf{Public datasets.}
We incorporate multiple benchmark datasets spanning the four domains (see Table~\ref{tab:dataset_summary} and supplementary materials for details). Despite their usefulness, these datasets exhibit imbalance, such as the scarcity of large-scale flare removal data and limited diversity in haze scenarios.

\noindent\textbf{Data expansion.}
To remedy these gaps and increase data volume, we expand training data through three sources: real-world captures, 2D synthesis, and 3D rendering.
\begin{itemize}[nosep,leftmargin=*]
\item \textit{Lens flare.} The key bottleneck lies in insufficient data. We therefore construct 78 indoor and outdoor 3D scenes in Blender~\citep{blender2018blender}, rendering about 70K paired images, named HALO dataset. Unlike Flare7K~\citep{dai2022flare7k}, which overlays flare layers on clean images, our rendered data produce physically consistent and realistic flare effects. The dataset covers diverse flare patterns, including reflective flare, glare, shimmer, and streaks.
\item \textit{Shadow.} While public datasets cover both indoor and outdoor scenes, they contain only $\sim$5K pairs. To scale up, we add an additional 26K photo pairs. Specifically, we repurpose internal object-effect removal data: by stitching objects without shadows into background images, we synthesize corresponding shadow-free versions to form the Large Real-world Shadow Removal Dataset (LR-SRD).
\item \textit{Haze.} Existing synthetic datasets (RESIDE, HAZESPACE) often appear uniform or algorithmically simplistic. To generate more realistic and challenging cases, we use their clean ground-truth images with monocular depth~\citep{ke2025marigold}, and apply a physically motivated atmospheric rendering pipeline. This allows precise control of parameters such as visibility, airlight color, scatter, and optical thickness. To simulate non-homogeneous haze or fog, we introduce procedural noise fields and path blurring, yielding realistic textures of haze, smoke, and fog. More synthesis details are provided in the supplementary material.
\end{itemize}
These expanded datasets extend coverage to underrepresented scenarios, enhancing UniSER’s robustness in the wild. A detailed breakdown is given in Table~\ref{tab:dataset_summary}, with representative samples in Fig.~\ref{fig:dataset_vis}.

\begin{figure*}[t]
  \centering
   \includegraphics[width=0.95\linewidth]{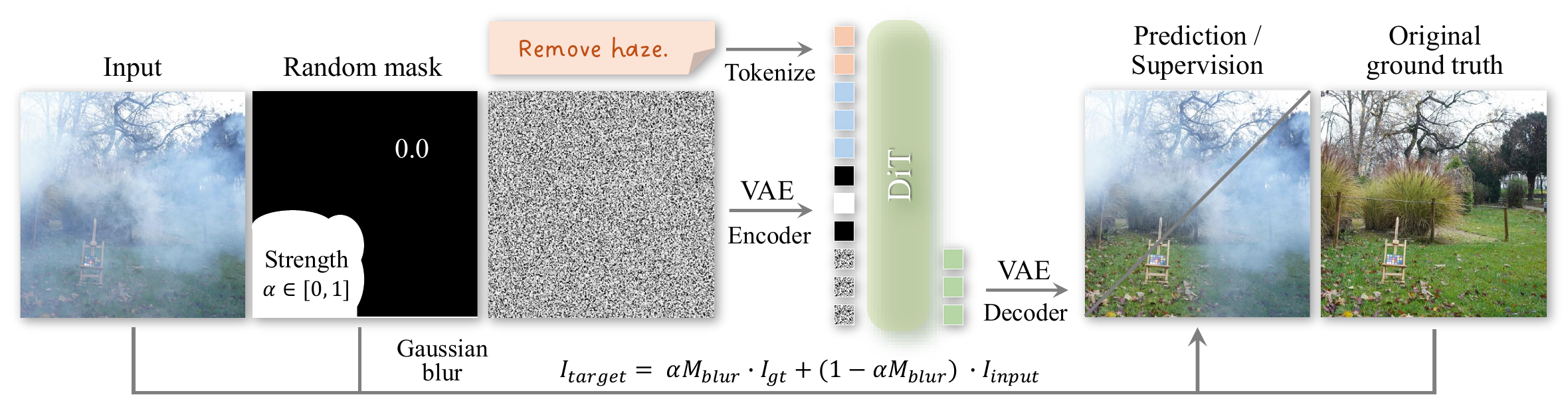}
   \vspace{-4mm}
   \caption{The architecture of UniSER. During training, the mask is randomly synthesized along with a scalar strength, and the supervision  is composed by the input image and the original ground truth via the mask and the strength. } 
   \vspace{-3mm}
   \label{fig:main_arch}
\end{figure*}

\subsection{Framework}

As shown in Fig.~\ref{fig:main_arch}, UniSER is a unified framework designed to tackle multiple soft effect removal tasks. Inspired by UniReal~\citep{chen2025unireal}, the core architecture reformulates these diverse tasks as a problem of \textit{discontinuous frame generation} within a latent diffusion model. The process begins with a Variational Autoencoder (VAE)~\citep{kingma2013auto} encoding the input image into a compact latent space, while a text encoder processes a task-specific prompt (\textit{e.g.}, ``\textit{remove haze}'') to generate instructive embeddings. These conditional inputs (image latent and textual embeddings) are then concatenated with the noisy target latent and fed as a sequence to a Diffusion Transformer (DiT). The DiT's full attention mechanism operates on this sequence, allowing it to iteratively predict and remove noise from the target latent by conditioning on both the visual context and the textual instructions. Finally, the fully denoised latent is passed through the VAE decoder to reconstruct the final, effect-free image. The model is trained using a mean squared error (MSE) loss between the predicted noise and the ground truth noise, with a timestep-dependent weighting scheme to balance the contributions of different noise levels.

\noindent\textbf{Random Masking Strategy.}
As established in the framework, a mask can be supplied as a condition to guide the denoising process toward a specific spatial region. However, most of the training sets do not contain the mask of effects. To ensure the model can robustly handle any user-provided mask shape, we adopt a random masking strategy. During training, following~\citep{suvorov2022resolution,zheng2022cm} we synthesize a wide variety of binary masks $M$ by randomly combining geometric primitives like rectangles with free-form, stroke-like patterns that simulate user brush strokes. Afterwards, providing pairs $\{I_{input}, I_{gt}\}$, we generate the corresponding training supervision ${I_{target}}$ where the effect is removed only within the masked region via simply compositing  $I_{input}$ and $I_{gt}$ with the mask, as shown in Equation~\ref{eq:composition}. Note that the regions of effects in the input image are unavailable, hence the masks do not necessarily cover them. In this way the behaviors the model to learn is summarized as following:
\begin{itemize}[nosep,leftmargin=*]
    \item Region inside the mask w/ effects: remove effects based on the strength;
    \item Region inside the mask w/o effects: keep identical;
    \item Region outside the mask: keep identical.
\end{itemize}
Additionally, to make the supervision natural-looking, we blur the mask boundary via dilation and Gaussian blur. 
This strategy exposes the model to a vast distribution of possible mask shapes, enhancing its generalization capability for arbitrary user edits, and removing the sepcific effect regions.

\noindent\textbf{Removal Strength Control.}
Beyond specifying \textit{where} to remove an effect, UniSER allows users to control \textit{how much} of the effect is removed. This is achieved by training the model to interpret continuous values in the conditional mask as an indicator of removal intensity. During the training process, for each sample, we uniformly sample a floating-point scalar value to represent ``strength'', denoted as $\alpha\in[0, 1]$. Instead of conditioning the model on a binary mask $M$, we provide a soft value mask $ \alpha M$. The model thus learns to associate a mask value of 1.0 with complete removal, 0.0 with no change, and intermediate values with partial removal. 
On the other hand, along with the aforementioned blurred mask, the training target is generated by linearly interpolating between the clean ground truth ($I_{gt}$) and the input with effects ($I_{input}$) using the randomly sampled $\alpha$. 
Formally, the supervision during training is computed as following:
\vspace{-1mm}
\begin{equation}
    I_{target} = \alpha M_{blur} \cdot I_{gt} + (1 - \alpha M_{blur}) \cdot I_{input}
    \label{eq:composition}
\end{equation}
This joint strategy of conditioning on a soft mask while generating a correspondingly blended target enables the model to learn a continuous and intuitive mapping from the control signal to the desired degree of effect removal. 

\noindent\textbf{Handling Undefined Effects.}
Our framework also extends to zero-shot generalization on unseen soft effects through two complementary fine-tuning strategies. First, we randomly replace task-specific prompts with a generic prompt \textit{``remove effects''}, encouraging the model to capture a shared notion of removal across tasks. Second, we introduce an auxiliary task using clean images: random masks are generated and overlaid with semi-transparent or opaque regions to synthesize degraded inputs, which are trained exclusively with the generic prompt. This prevents overfitting to predefined effect categories and compels the model to learn the broader concept of removing arbitrary occlusions, thereby enabling generalized restoration.

\noindent\textbf{Adding \& Enhancing Effects.}
We can easily invert the removal task to adding or enhancing effects by swapping the roles of the input and the target. Similarly, the adding or enhancing ability is controlled by the mask and strength given by users. We demonstrate this ability in Fig.~\ref{fig:additional}.
\begin{figure*}[t]
  \centering
   \includegraphics[width=1.0\linewidth]{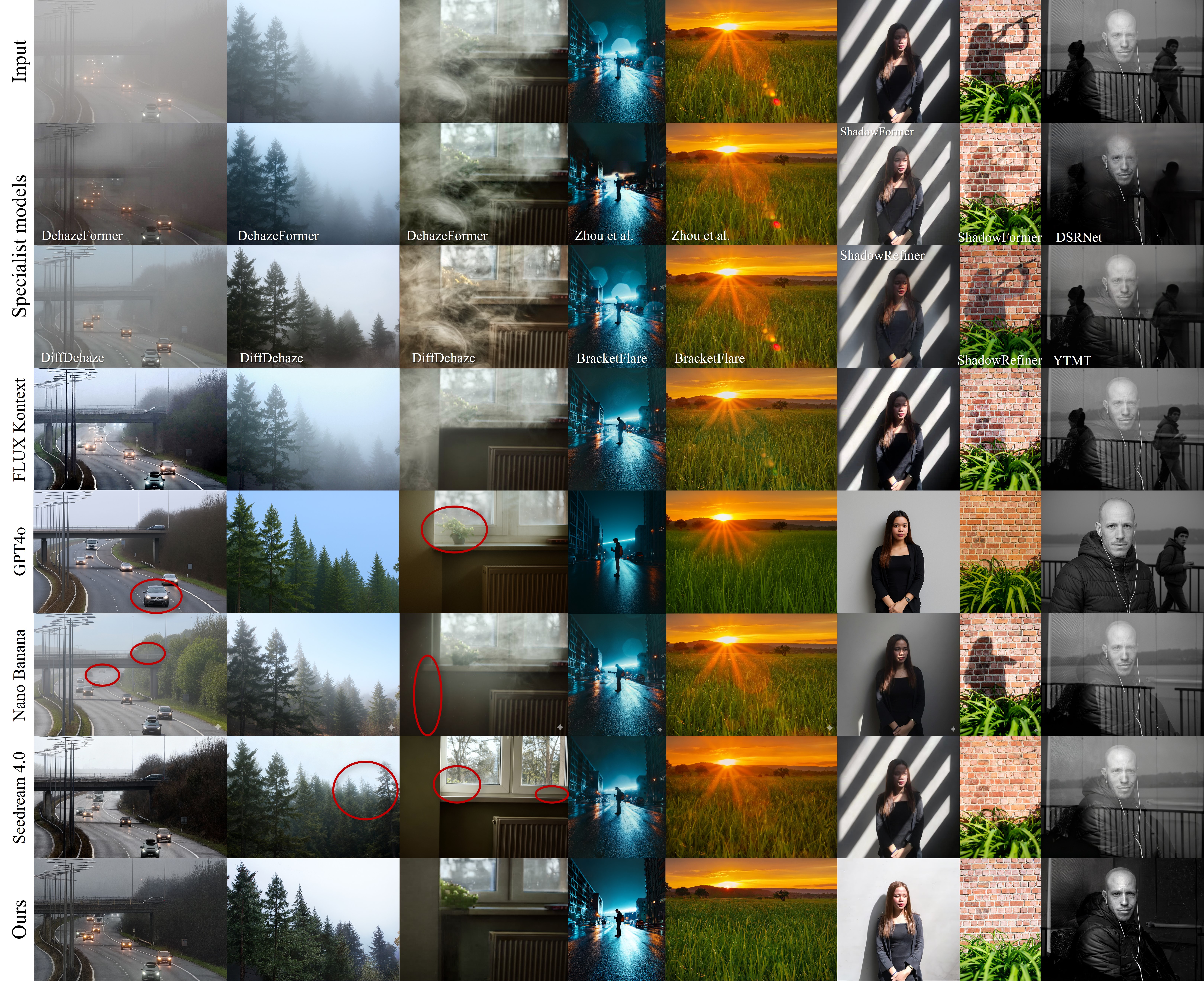}
   \vspace{-6mm}
   \caption{Comparisons with state-of-the-art specialist and generalist models on in-the-wild testing data. For effect removal, our method significantly outperforms these baselines. Moreover, generalist models fail to preserve the identity of background objects, some of the discrepancies are circled, better view by zooming in.}
   \vspace{-5mm}
   \label{fig:comp_sota}
\end{figure*}

\begin{table*}[t]
\centering
\caption{No-reference quantitative comparison on in-the-wild images for four SER tasks. We report results from multiple image quality assessment metrics.}
\vspace{-1mm}
\label{tab:comp_1_compact_no_musiq}
\setlength{\tabcolsep}{5.4mm} 
\scalebox{0.75}{
\begin{tabular}{lccc c lccc}
\toprule
\multicolumn{4}{c}{\textbf{Haze}} & & \multicolumn{4}{c}{\textbf{Shadow}} \\
\cmidrule(lr){1-4} \cmidrule(lr){6-9}
\textbf{Method} & \textbf{LIQE$\uparrow$} & \textbf{Contrast$\uparrow$} & \textbf{QwenQA$\uparrow$} & & \textbf{Method} & \textbf{LIQE$\uparrow$} & \textbf{Contrast$\uparrow$} & \textbf{QwenQA$\uparrow$} \\
\midrule
Dehazeformer & 1.9999 & +0.74 & 0.0 & & ShadowFormer & 3.3704 & +3.09 & 18.8 \\
DiffDehaze & 1.5624 & +0.03 & 9.1 & & ShadowRefiner & 3.5179 & -2.30 & 26.3 \\
Flux Kontext & 2.2584 & +3.85 & 22.7 & & Flux Kontext & 3.3184 & +0.73 & 36.3 \\
Nano Banana & 2.6864 & +0.26 & 27.3 & & Nano Banana & 3.6399 & -4.93 & 35.0 \\
Seedream 4.0 & 2.1253 & +2.60 & 52.7 & & Seedream 4.0 & 2.7640 & -3.58 & 36.3 \\
\textbf{Ours} & \textbf{2.8225} & \textbf{+5.57} & \textbf{60.0} & & \textbf{Ours} & \textbf{3.7764} & \textbf{+3.61} & \textbf{65.0} \\
\midrule
\multicolumn{4}{c}{\textbf{Lens Flares}} & & \multicolumn{4}{c}{\textbf{Reflections}} \\
\cmidrule(lr){1-4} \cmidrule(lr){6-9}
\textbf{Method} & \textbf{LIQE$\uparrow$} & \textbf{Contrast$\uparrow$} & \textbf{QwenQA$\uparrow$} & & \textbf{Method} & \textbf{LIQE$\uparrow$} & \textbf{Contrast$\uparrow$} & \textbf{QwenQA$\uparrow$} \\
\midrule
Uformer & 1.3832 & -4.39 & 30.9 & & YTMT & 1.1187 & -2.30 & 14.4 \\
BracketFlare & 3.3377 & -10.72 & 13.6 & & DSRNet & 1.6975 & -4.17 & 17.8 \\
Flux Kontext & 3.0574 & -0.31 & 62.7 & & Flux Kontext & 1.7009 & +0.71 & 8.9 \\
Nano Banana & 3.0358 & -4.05 & 71.8 & & Nano Banana & 2.0935 & -1.25 & 56.7 \\
Seedream 4.0 & 2.1643 & -4.41 & 73.6 & & Seedream 4.0 & 1.6145 & -0.96 & 53.5 \\
\textbf{Ours} & \textbf{3.5186} & \textbf{+2.33} & \textbf{92.7} & & \textbf{Ours} & \textbf{2.2257} & \textbf{+1.83} & \textbf{75.6} \\
\bottomrule
\end{tabular}
}
\vspace{-3mm}
\end{table*}

\begin{table*}[t]
\centering
\caption{Quantitative comparison with state-of-the-art methods across four soft effect removal tasks. We report PSNR ($\uparrow$) and SSIM ($\uparrow$) on eight benchmarks. Our unified model is compared against specialist methods in each respective category.}
\label{tab:comp_2}
\vspace{-1mm}
{ 
\setlength{\tabcolsep}{6.4mm} 
\scalebox{0.75}{
\begin{tabular}{lcc cc lcccc}
\toprule
\multicolumn{3}{c}{\textbf{Lens Flares}} & & & \multicolumn{5}{c}{\textbf{Haze}} \\
\cmidrule(lr){1-3} \cmidrule(lr){6-10}
\textbf{Method} & \multicolumn{2}{c}{Flare7k} & & & \textbf{Method} & \multicolumn{2}{c}{HSTS} & \multicolumn{2}{c}{SOTS} \\
\cmidrule(lr){2-3} \cmidrule(lr){7-8} \cmidrule(lr){9-10}
& PSNR & SSIM & & & & PSNR & SSIM & PSNR & SSIM \\
\midrule
\textit{Zhang et al.}~\cite{zhang2020nighttime} & 21.02 & 0.784 & & & DCP & 17.01 & 0.803 & 18.38 & 0.819 \\
\textit{Zhou et al.}~\cite{zhou2023improving} & 25.18 & 0.872 & & & AOD-Net & 19.68 & 0.835 & 20.08 & 0.861 \\
UNet~\citep{dai2022flare7k} & 26.11 & 0.879 & & & GCANet & 21.37 & 0.874 & 21.66 & 0.867 \\
Restormer~\citep{dai2022flare7k} & 26.28 & 0.883 & & & PSD & 19.37 & 0.824 & 20.49 & 0.844 \\
Uformer~\citep{dai2022flare7k} & 26.98 & 0.890 & & & MSFNet & 31.03 & 0.931 & \textbf{30.07} & 0.939 \\
Difflare & 26.06 & \textbf{0.898} & & & UCL-Dehaze & 26.87 & 0.933 & 25.21 & 0.927 \\
\midrule
\textbf{Ours} & \textbf{27.34} & {0.891} & &  & \textbf{Ours} & \textbf{32.17} & \textbf{0.962} & {29.52} & \textbf{0.955} \\
\bottomrule
\end{tabular}
}
} 


{ 
\setlength{\tabcolsep}{3.3mm} 
\scalebox{0.75}{
\begin{tabular}{lcccccc c lcccc}
\toprule
\multicolumn{7}{c}{\textbf{Shadow}} & & \multicolumn{5}{c}{\textbf{Reflections}} \\
\cmidrule(lr){1-7} \cmidrule(lr){9-13}
\textbf{Method} & \multicolumn{2}{c}{WSRD+} & \multicolumn{2}{c}{ISTD+} & \multicolumn{2}{c}{SRD} & & \textbf{Method} & \multicolumn{2}{c}{SIR2} & \multicolumn{2}{c}{Nature20} \\
\cmidrule(lr){2-3} \cmidrule(lr){4-5} \cmidrule(lr){6-7} \cmidrule(lr){10-11} \cmidrule(lr){12-13}
& PSNR & SSIM & PSNR & SSIM & PSNR & SSIM & & & PSNR & SSIM & PSNR & SSIM \\
\midrule
ShadowFormer & 25.44 & 0.820 & 32.78 & 0.934 & 30.58 & 0.958 & & \textit{Zhang et al.}~\cite{zhang2018single} & 22.45 & 0.872 & 20.37 & 0.772 \\
ShadowRefiner & 26.04 & 0.827 & 31.03 & 0.928 & - & - & & YTMT & 23.05 & 0.886 & 21.03 & 0.802 \\
DCShadowNet & 21.62 & 0.593 & 25.50 & 0.694 & - & - & & DSRNet & 24.97 & 0.907 & 21.70 & 0.820 \\
ShadowDiffusion & - & - & 31.08 & 0.950 & 31.91 & 0.968 & & PromptRR & 24.22 & 0.876 & 21.00 & 0.814 \\
StableShadowDiff & 26.26 & 0.827 & 35.19 & \textbf{0.970} & 33.63 & 0.968 & & L-DiffER & 25.18 & 0.911 & 23.95 & \textbf{0.831} \\
\midrule
\textbf{Ours} & \textbf{26.91} & \textbf{0.829} & \textbf{35.59} & {0.964} & \textbf{34.16} & \textbf{0.971} & & \textbf{Ours} & \textbf{25.98} & \textbf{0.911} & \textbf{24.17} & {0.812} \\
\bottomrule
\end{tabular}
}
} 
\vspace{-4mm}
\end{table*}

\section{Experiments}

\subsection{Benchmarks and Baselines}
\noindent\textbf{Benchmarks.}
We evaluate UniSER across four soft-effect tasks on widely used benchmarks. For \textit{lens flare removal}, we adopt the Flare7K real-world test set~\citep{dai2022flare7k}. For \textit{shadow removal}, we test on SRD~\citep{qu2017deshadownet}, ISTD+~\citep{wang2018stacked}, and the high-resolution WSRD+~\citep{vasluianu2023wsrd}. For \textit{haze removal}, we use the SOTS and HSTS subsets of RESIDE~\citep{li2018benchmarking}. For \textit{reflection removal}, we employ $SIR^2$~\citep{wan2017benchmarking} and the Nature test set~\citep{li2020single}. UniSER is fine-tuned on the training splits of these datasets for domain adaptation. Evaluation uses standard full-reference metrics: PSNR and SSIM.

To assess real-world robustness, we collected 39 in-the-wild images containing haze, fog, flare, reflection, and shadow. As no ground truth is available, we report reference-free metrics (LIQE~\citep{zhang2023blind}, contrast gain~\citep{wang2024ucl}), and a reference-based evaluation with Qwen2.5-VL-72B~\citep{bai2025qwen2}, a vision-language model instructed to judge the percentage of effect removal. We will further discuss these metrics in the supplementary material.

\noindent\textbf{Baselines.}
We compare against both generalist and specialist methods. Generalist baselines include GPT-4o~\citep{hurst2024gpt}, FLUX Kontext~\citep{labs2025flux}, Nano Banana~\citep{google:gemini2.5flash_image}, and Seedream 4.0~\citep{bytedance:seedream4}. Specialist baselines cover:

\noindent\textit{Lens flare}:~\citep{zhang2020nighttime,zhou2023improving,dai2022flare7k}, BracketFlare~\citep{dai2023nighttime}, Difflare~\citep{zhou2024difflare};

\noindent\textit{Dehazing}: DCP~\citep{he2010single}, AOD-Net~\citep{li2017aod}, GCANett~\citep{chen2019gated}, PSD~\citep{chen2021psd}, Dehazeformer~\citep{song2023vision}, MSF-Net~\citep{zhu2021multi}, UCL-Dehazet~\citep{wang2024ucl}, DiffDehaze~\citep{wang2025learning};

\noindent\textit{Shadow removal}: ShadowFormer~\citep{guo2023shadowformer}, ShadowRefiner~\citep{dong2024shadowrefiner}, DCShadowNet~\citep{jin2021dc}, ShadowDiffusion~\citep{guo2023shadowdiffusion}, StableShadowDiff~\citep{xu2025detail};

\noindent\textit{Reflection removal}:~\citep{zhang2018single}, YTMT~\citep{hu2021trash}, DSRNet~\citep{hu2023single}, PromptRR~\citep{wang2024promptrr}, L-DiffER~\citep{hong2024differ}.

\subsection{Comparisons with State-of-The-Art}

\noindent\textbf{Qualitative Comparisons.} Fig.~\ref{fig:comp_sota} visually compares UniSER with state-of-the-art models on challenging in-the-wild images. Specialist models generalize poorly to out-of-domain data, often resulting in incomplete removal or new artifacts. Meanwhile, powerful generalist models like Nano Banana and FLUX Kontext suffer from instability and fail to preserve scene details, leading to significant content drift (highlighted by red circles). In contrast, UniSER effectively removes a wide range of soft effects while remaining highly faithful to the original image content, producing clean and content-consistent results.

\noindent\textbf{Quantitative Comparisons.} To assess real-world generalization,we first conduct a comparison on a challenging in-the-wild test set using no-reference metrics, shown in Table~\ref{tab:comp_1_compact_no_musiq}. In this more difficult setting, UniSER significantly outperforms both specialist and generalist baselines in terms of perceptual quality and removal efficacy, achieving the highest LIQE, Contrast gain, and QwenQA scores across nearly all tasks, which highlights its robust generalization. We then evaluate UniSER against specialists on eight standard benchmarks using full-reference metrics (Table~\ref{tab:comp_2}). The results show our unified model achieves state-of-the-art performance, consistently outperforming or matching specialist models by obtaining top scores across all four tasks, including the highest PSNR on multiple benchmarks.

\begin{figure*}[t]
  \centering
   \includegraphics[width=0.95\linewidth]{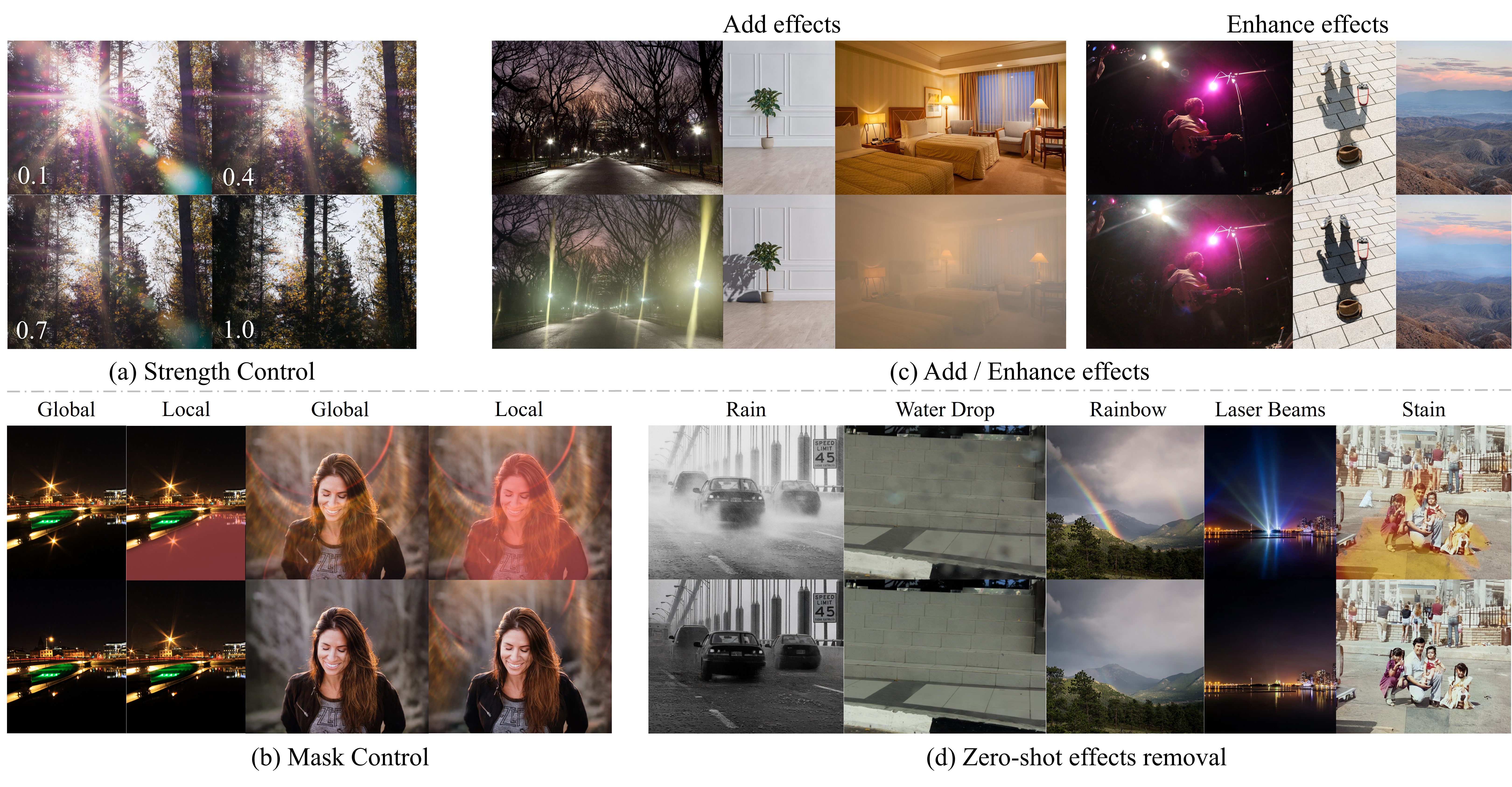}
   \vspace{-5mm}
   \caption{(a) Illustration of Strength Control for effect removal. (b) Illustration of Mask Control for accurate user regional editing. (c) Adding realistic effects to clean image, or enhance current effects for flexible editing purpose. (d) Zero-shot generalization ability on multiple unseen degradations like rain, stain, etc.}
   \vspace{-4mm}
   \label{fig:additional}
\end{figure*}

\subsection{Ablations and Applications}
\vspace{-1mm}
\noindent\textbf{Joint effect removal.}
We conduct an ablation study to validate the effectiveness of our joint-task learning strategy. As shown in Table~\ref{tab:ablation_mtl_stl}, we compare our full model, trained with Joint-Task Learning (JTL), against four same models trained independently using Single-Task Learning (STL). The results clearly indicate that the JTL model consistently outperforms the STL models across all four tasks on their respective benchmarks. This superiority suggests that by learning a unified representation from diverse soft effects, UniSER develops a more robust and generalizable feature space that benefits all individual tasks.

\noindent\textbf{Strength control.}
As illustrated in Figure~\ref{fig:additional}(a), UniSER provides fine-grained control over the intensity of the effect removal. Users can specify a continuous strength value, allowing for a smooth transition from partial reduction to complete effect removal. This feature offers greater flexibility for users to achieve their desired level of restoration.

\noindent\textbf{Mask control.}
UniSER supports precise, localized editing through mask-based control, as shown in Figure~\ref{fig:additional}(b). By providing a binary mask, users can designate specific spatial regions for effect removal while leaving the rest of the image untouched. This allows for targeted and accurate edits tailored to user needs.

\noindent\textbf{Effects addition and enhancement.}
Beyond removal, the UniSER framework is also capable of generative tasks. As demonstrated in Figure~\ref{fig:additional}(c), by inverting the process, our model can realistically add new soft effects to clean images or enhance existing ones. This versatility makes it a valuable tool for creative editing and data augmentation.

\noindent\textbf{Zero-shot removal.}
UniSER exhibits strong generalization capabilities to novel degradations not seen during training. As shown in Figure~\ref{fig:additional}(d), the model can perform zero-shot removal of unseen artifacts such as rain and stains. This ability underscores the robustness of the learned features and the model's potential to handle a wider range of image restorations beyond its core training tasks.

\noindent\textbf{Reproducibility Statement}
The portion of our method that relies on public datasets is reproducible, as our implementation is based on the open-source DiT codebase, we will release an open-source version of UniSER upon acceptance.

\begin{table}[t]
\centering
\caption{Ablation study on training strategies. JTL (Joint-Task Learning) represents our full UniSER, while STL (Single-Task Learning) denotes models trained separately for each task.}
\vspace{-3mm}
\label{tab:ablation_mtl_stl}
\setlength{\tabcolsep}{0.8mm} 
\scalebox{0.75}{
\begin{tabular}{lcccccccc}
\toprule
& \multicolumn{2}{c}{\textbf{Lens Flares}} & \multicolumn{2}{c}{\textbf{Haze}} & \multicolumn{2}{c}{\textbf{Shadow}} & \multicolumn{2}{c}{\textbf{Reflections}} \\
\cmidrule(lr){2-3} \cmidrule(lr){4-5} \cmidrule(lr){6-7} \cmidrule(lr){8-9}
\textbf{Method} & \multicolumn{2}{c}{Flare7k} & \multicolumn{2}{c}{HSTS} & \multicolumn{2}{c}{ISTD+} & \multicolumn{2}{c}{SIR2-wild} \\
\cmidrule(lr){2-3} \cmidrule(lr){4-5} \cmidrule(lr){6-7} \cmidrule(lr){8-9}
& PSNR$\uparrow$ & SSIM$\uparrow$ & PSNR$\uparrow$ & SSIM$\uparrow$ & PSNR$\uparrow$ & SSIM$\uparrow$ & PSNR$\uparrow$ & SSIM$\uparrow$ \\
\midrule
STL & 27.18 & 0.890 & 31.91 & \textbf{0.963} & 35.43 & {0.963} & 26.40 & 0.876 \\
JTL & \textbf{27.34} & \textbf{0.891} & \textbf{32.17} & {0.962} & \textbf{35.59} & \textbf{0.964} & \textbf{27.44} & \textbf{0.918} \\
\bottomrule
\end{tabular}
}
\end{table}

\section{Conclusion and Limitations}
\vspace{-1mm}
We introduced UniSER, a unified foundation model that validates a data-centric methodology for Soft Effects Removal (SER) task, which effectively handles diverse degradations including lens flare, haze, shadows, and reflections. By curating a large-scale dataset with hugh-quality pairs and training with dedicated controls, UniSER overcomes the poor generalization of specialist models and the content inconsistency of generalist approaches. Extensive experiments demonstrate that our model achieves state-of-the-art performance on standard benchmarks and superior perceptual quality on in-the-wild images while providing fine-grained user controls, supports creative effect generation, and shows strong zero-shot generalize capabilities. Key limitations include its significant computational cost and the extensive resources required for training. Nevertheless, UniSER represents a significant step towards a universal and controllable solution for high-fidelity image restoration.
{
    \small
    \bibliographystyle{ieeenat_fullname}
    \bibliography{main}
}

\clearpage
\setcounter{page}{1}
\maketitlesupplementary

In this supplementary material, we are going to illustrate i) more details of our data curation details; ii) more details of the haze synthetic pipeline; iii) the detail design of non-reference metrics; iv) more implementation details and v) more visual results and quality analysis.

\section{Data Curation on Public Datasets.}

Our data collection process aggregates established benchmarks from each domain. For lens flare removal, we incorporate the real-world paired dataset FlareReal600~\citep{dai2024mipi} for nighttime optical artifacts. For shadow removal, our dataset combines several widely-used benchmarks, including SRD~\citep{qu2017deshadownet}, ISTD+~\citep{wang2018stacked}, and the high-resolution WSRD+~\citep{vasluianu2023wsrd}, to cover a wide variety of shadow types and complexities. The most extensive category is haze removal, for which we collected a diverse range of datasets. This includes smaller, real-world datasets captured under controlled conditions, we name this set as Haze-R, including: I-HAZE~\citep{ancuti2018ihaze}, O-HAZE~\citep{ancuti2018ohaze}, Dense-Haze~\citep{ancuti2019dense}, NH-Haze~\citep{ancuti2020nh,ancuti2021ntire,ancuti2023ntire,ancuti2024ntire}, and video dehaze dataset REVIDE~\citep{zhang2021learning}, multi-level haze dataset LM-Haze~\citep{zhang2024lmhaze}. Large-scale synthetic datasets that provide broad coverage of different haze conditions like RESIDE~\citep{li2018benchmarking} and HAZESPACE2M~\citep{islam2024hazespace2m} are also included. Finally, for reflection removal, we integrated datasets that capture various scenarios, such as general real-world reflections RRW~\citep{zhu2024revisiting}, polarization-based captures POLAR-RR~\citep{lei2020polarized}, and flash-induced reflections RFC~\citep{lei2021robust}, and synthetic by overlaying dataset BDN~\citep{yang2018seeing}.
However, these publicly available datasets were originally collected for specific tasks. As a result, their overall distribution is imbalanced, including discrepancies across different tasks, between real and synthetic data, as well as between indoor and outdoor scenes, and day and night conditions.

\section{Details of the Haze Synthesis Pipeline}
 A significant portion of our training dataset, particularly for atmospheric effects like haze, fog, and smoke, was generated using a custom synthesis pipeline. This pipeline was designed to overcome the limitations of existing synthetic datasets, which often lack physical realism and diversity. Our methodology is built upon two core components: (1) a physically-motivated atmospheric rendering engine that applies uniform atmospheric effects based on scene geometry, and (2) a procedural texture generator that creates complex, non-homogeneous patterns to simulate phenomena like patchy fog or smoke plumes.

\subsection{Physically-Motivated Atmospheric Rendering Model}
The foundation of our synthesis pipeline is a unified rendering model inspired by the Radiative Transfer Equation (RTE). This model mathematically describes how light interacts with a participating medium (like haze or fog) as it travels from a scene object to the camera. The final color at a pixel $x$, denoted $I_{out,c}(x)$ for a color channel $c$, is a composite of the attenuated scene radiance and the in-scattered light from the atmosphere, known as airlight.

The image formation model is expressed as:
\begin{equation}
    I_{out,c}(x) = I_{in,c}(x) \cdot T_c(x) + A_c \cdot (\omega_{0,c} \cdot \kappa) \cdot (1 - T_c(x)^\eta)
\end{equation}
where:
\begin{itemize}
    \item $I_{in,c}(x)$ is the original, effect-free color of the scene at pixel $x$.
    \item $T_c(x)$ is the \textbf{transmittance}, representing the fraction of light that successfully travels from the object to the camera without being scattered or absorbed.
    \item $A_c$ is the color of the \textbf{airlight}, which is the ambient environmental light scattered towards the camera by the atmospheric particles. This parameter is crucial for defining the hue of the haze (e.g., white for fog, sky-tinted for haze, warm gray for smoke).
    \item $\omega_{0,c}$ is the \textbf{single-scattering albedo}, a value in $[0, 1]$ indicating the proportion of light extinction that is due to scattering versus absorption. For non-absorptive media like fog and haze, $\omega_0 \approx 1.0$. For absorptive media like smoke, $\omega_0 < 1.0$.
    \item $\kappa$ is an \textbf{anisotropy gain factor}, derived from the Henyey-Greenstein phase function. It accounts for directionality of scattering (i.e., whether particles scatter light more strongly forward or backward). For simplicity in our large-scale synthesis, we set $\kappa=1$, modeling isotropic scattering.
    \item $\eta$ is a \textbf{multiple-scattering boost exponent} $(0 < \eta \le 1)$. This term provides a compact approximation for the effects of multiple scattering events. A lower value of $\eta$ increases the brightness of the veil, simulating the appearance of denser media where light scatters multiple times before reaching the camera.
\end{itemize}

\begin{figure*}[t]
  \centering
   \includegraphics[width=0.85\linewidth]{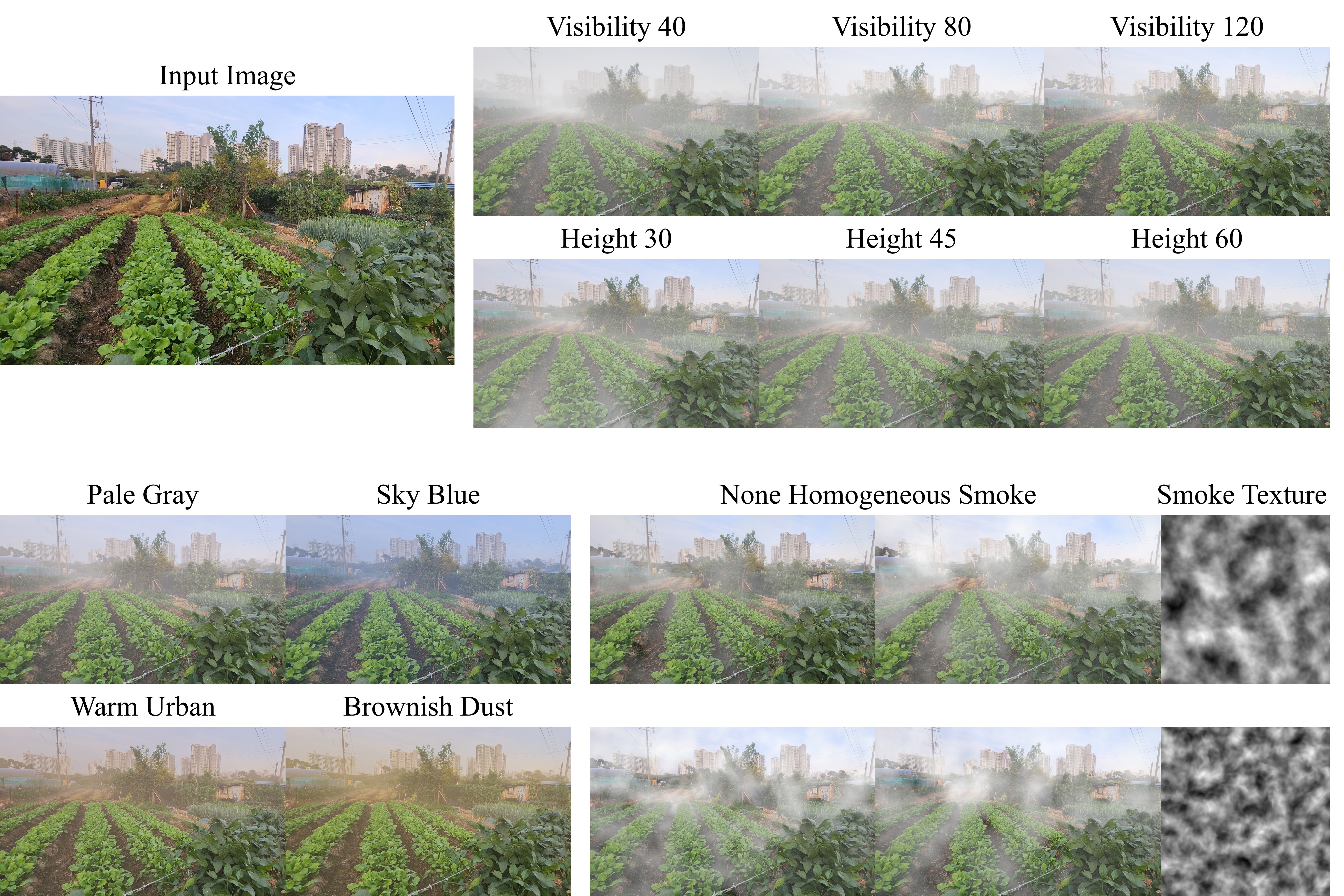}
   \vspace{-2mm}
   \caption{Visualization of our synthetic haze generated by our the proposed pipeline. Our method is capable of synthesizing multiple essences of haze, fog and smoke, within different colors, morphologies and optical properties. } 
   \label{fig:haze_vis}
\end{figure*}

\subsubsection{Optical Depth and Transmittance}

The transmittance $T_c(x)$ is determined by the optical depth $\tau_c(x)$ of the medium along the line of sight, following the Beer-Lambert law:
\begin{equation}
    T_c(x) = e^{-\tau_c(x)}
\end{equation}
The optical depth is the integral of the extinction coefficient $\beta_{t,c}$ over the distance $d(x)$ from the camera to the object at pixel $x$. To model realistic atmospheres, we assume an exponential decay of particle density with height $h$:
\begin{equation}
    \beta_{t,c}(h) = \beta_{t0,c} \cdot e^{-h/H}
\end{equation}
where $\beta_{t0,c}$ is the base extinction coefficient at a reference height (e.g., sea level), and $H$ is the \textbf{scale height}, which defines how rapidly the atmosphere thins out. For a near-horizontal viewing angle, the optical depth can be approximated as:
\begin{equation}
    \tau_c(x) \approx \beta_{t0,c} \cdot e^{-h(x)/H} \cdot d(x)
\end{equation}
The base extinction coefficient $\beta_{t0,c}$ is directly related to the meteorological visibility $V$ by the Koschmieder formula, $\beta_{t0} \approx 3.912 / V$.

\subsubsection{Geometric Inputs: Depth and Height}

Our rendering pipeline requires per-pixel geometric information.
\begin{itemize}[nosep,leftmargin=*]
    \item \textbf{Depth:} We use monocular depth maps estimated from the clean input images by Marigold~\citep{ke2025marigold}. These normalized depth maps are converted to distance in meters, $d(x)$, using a scene-specific maximum distance $d_{max}$.
    \item \textbf{Height:} When a true height map is unavailable, we utilize a \textbf{screen-space height proxy}: $h(x) = h_{max} \cdot (1 - y_{norm})$, where $y_{norm}$ is the normalized vertical coordinate of the pixel (0 at the top, 1 at the bottom). This proxy effectively treats pixels near the horizon as being at a higher altitude, enabling the synthesis of effects like low-lying valley fog that is denser at the bottom of the image.
\end{itemize}

\subsubsection{Color Space and Parameterization}

All physical calculations are performed in a linear RGB color space to ensure correctness. Input images, which are typically encoded in sRGB, are first decoded to linear space. After the atmospheric effects are composed, the resulting linear image is encoded back to sRGB. For our large-scale data generation, we programmatically varied all key parameters—including \textit{visibility}, \textit{airlight} color, \textit{eta}, and \textit{H}—across wide, physically plausible ranges to generate a diverse set of training pairs. We also introduced a random baseline value to the optical thickness $\tau$ in each render to add further variety.

\subsection{Procedural Generation of Non-Homogeneous Media}

To simulate complex, turbulent atmospheric effects like patchy fog or smoke, we integrated a procedural texture generator into our pipeline. This process creates realistic, wispy patterns that are used to spatially modulate the density of the rendered haze.

The generation process involves two main steps:
\begin{enumerate}
    \item \textbf{Vector Field Generation:} We first generate a 2D vector field $\vec{V}(\vec{p})$ for each pixel coordinate $\vec{p}=(x,y)$. The components of this field are determined by two independent layers of Perlin noise, $P(\cdot)$, distinguished by unique seeds ($\theta_1, \theta_2$), which simulates a turbulent flow field. The resulting vectors are normalized to create a unit vector field $\hat{V}(\vec{p})$:
    \begin{equation}
        \vec{V}(\vec{p}) = \begin{bmatrix} P(\vec{p}; \theta_1) \\ P(\vec{p}; \theta_2) \end{bmatrix}, \quad \hat{V}(\vec{p}) = \frac{\vec{V}(\vec{p})}{\| \vec{V}(\vec{p}) \| + \epsilon}
    \end{equation}
    where $\epsilon$ is a small constant to prevent division by zero.

    \item \textbf{Path Blurring (Advection):} A base noise texture, $M_0(\vec{p})$, is iteratively advected along the vector field $\hat{V}(\vec{p})$ for $N$ steps. In each step $k$, the new texture $M_{k+1}(\vec{p})$ is a blend of the previous texture $M_k(\vec{p})$ and a value sampled from a forward-projected position $\vec{p}'$. This technique smears the initial pattern, creating characteristic streaks. The update rule is:
    \begin{equation}
        M_{k+1}(\vec{p}) = (1-\alpha) \cdot M_k(\vec{p}) + \alpha \cdot M_k(\vec{p}')
    \end{equation}
    where $\vec{p}' = \vec{p} + \hat{V}(\vec{p}) \cdot \delta_s$. Here, $\delta_s$ is the step length, $\alpha$ is a blending factor (we use $\alpha=0.5$), and $M_k(\vec{p}')$ is obtained via bilinear interpolation as $\vec{p}'$ may have non-integer coordinates.
\end{enumerate}
The resulting grayscale texture after $N$ iterations, $M_N(\vec{p})$, is then used as a spatial density modulator, $M(x)$, for the extinction coefficient. The final optical depth calculation is modified to incorporate this texture:
\begin{equation}
    \tau_c(x) \approx (\beta_{t0,c} \cdot M(x)) \cdot e^{-h(x)/H} \cdot d(x)
\end{equation}
This allows us to render haze that is not uniform but varies in density and structure across the image, greatly enhancing the realism and challenge of our synthetic dataset. We also illustrate a sample image synthesized with multiple different types of haze, fog or smoke in Fig.~\ref{fig:haze_vis}.

\begin{figure*}[t]
  \centering
   \includegraphics[width=1.0\linewidth]{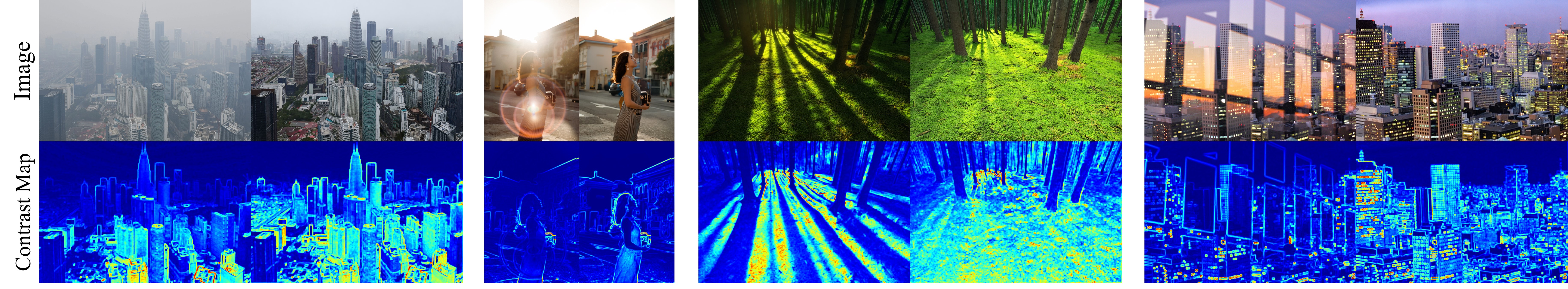}
   \vspace{-6mm}
   \caption{Contrast maps of image before and after edit by UniSER. Significant enhancements of contrast inside effect regions are observed, indicating our method successfully enhances the degraded image details.}
   \label{fig:additional}
\end{figure*}

\begin{figure*}[!ht]
  \centering
   \includegraphics[width=0.9\linewidth]{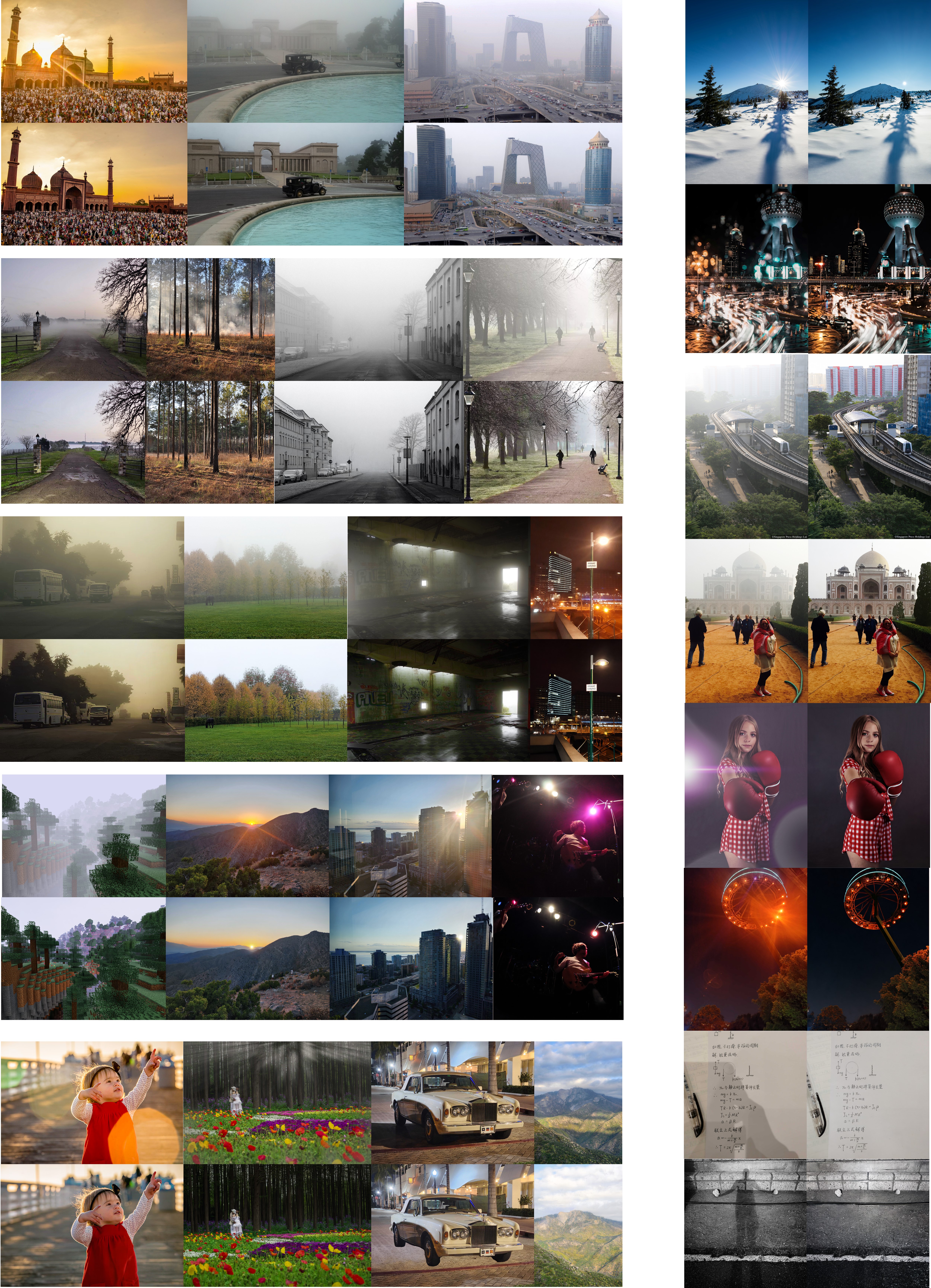}
   \vspace{-2mm}
   \caption{Gallery: Removing effects with UniSER. } 
   \label{fig:gallery_remove1}
\end{figure*}

\begin{figure*}[!ht]
  \centering
   \includegraphics[width=0.9\linewidth]{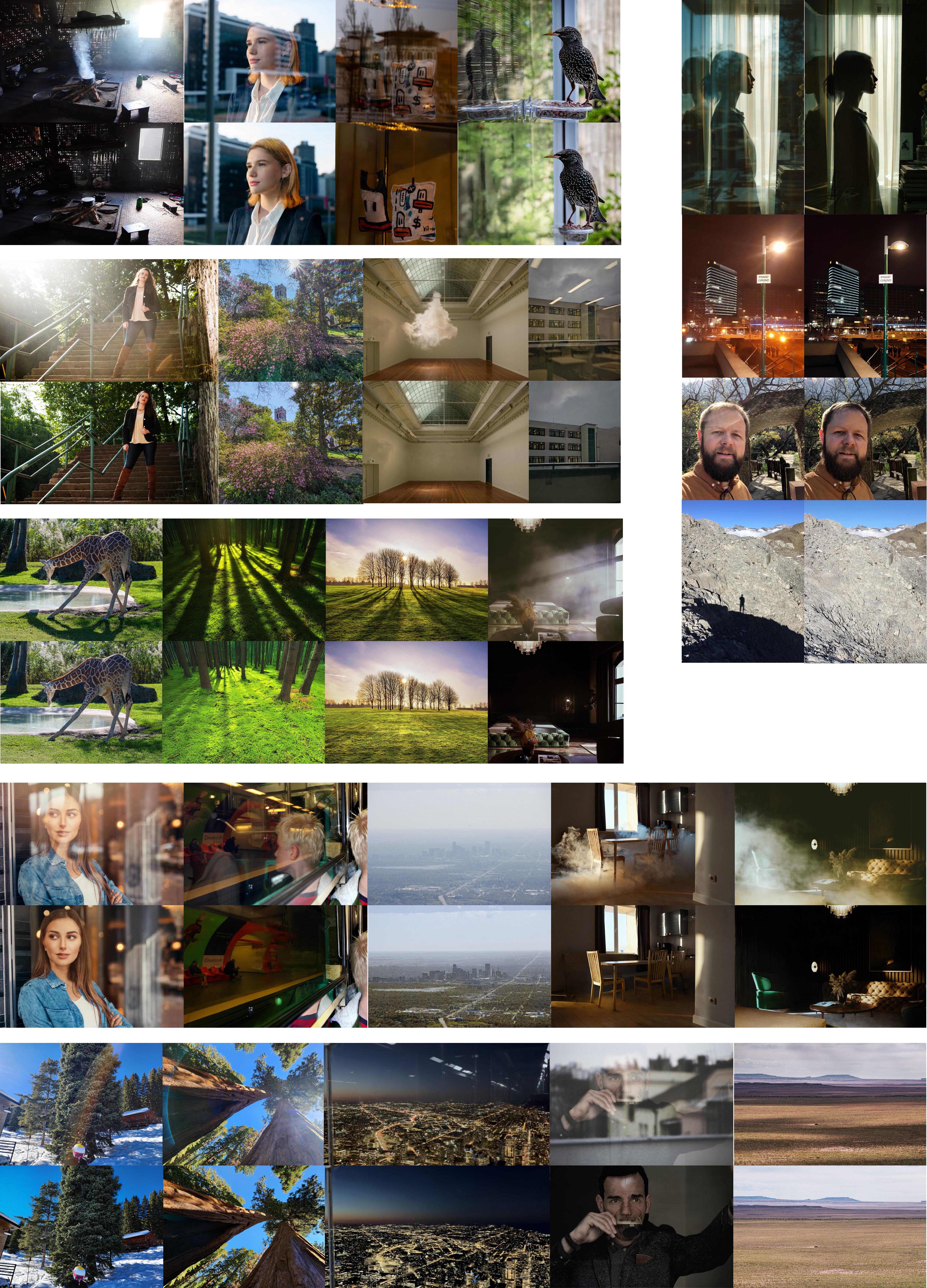}
   \vspace{-2mm}
   \caption{Gallery: Removing effects with UniSER. } 
   \label{fig:gallery_remove2}
\end{figure*}

\begin{figure*}[!ht]
  \centering
   \includegraphics[width=0.85\linewidth]{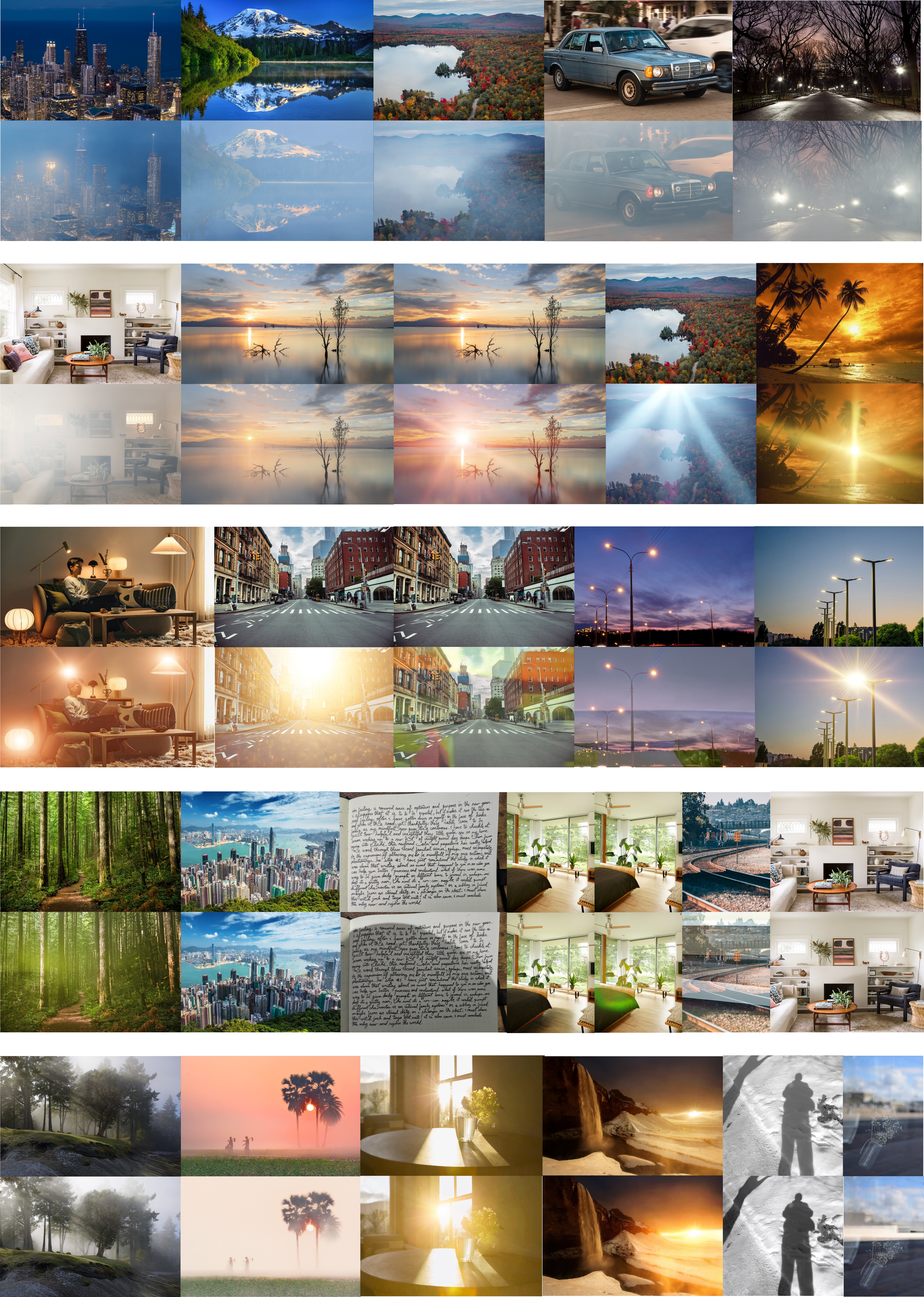}
   \vspace{-2mm}
   \caption{Gallery: Adding or enhancing effects with UniSER. } 
   \label{fig:gallery_add}
\end{figure*}

\section{Non-Reference Evaluation Metrics}
To rigorously assess the performance of our model on in-the-wild images where a ground-truth reference is unavailable, we employed specialized non-reference evaluation paradigms. These metrics are designed to provide both a quantitative measure of detail recovery and a qualitative score that emulates human perceptual judgment.

\subsection{Residual Contrast Gain}
While local contrast is a well-established indicator of image sharpness and detail, commonly iused in non-reference dehazing or similar tasks~\citep{wang2024ucl}. However since the measurements are averaged over the entire image, for localized effects like some types of lens flares or local shadows, the global evaluation is not significant. To overcome this limitation, we measure the \textbf{Residual Contrast Gain}, which quantifies the change in local contrast exclusively within the image regions modified by our model. This approach ensures that the evaluation focuses directly on the model's restoration efficacy. The computation is performed via the following steps:

\begin{enumerate}
    \item \textbf{Identification of Edited Regions.} Given a grayscale input image $I_{in}$ and the model's grayscale output $I_{out}$, we first identify the edited regions by computing a pixel-wise absolute difference map, $D(\vec{p}) = |I_{in}(\vec{p}) - I_{out}(\vec{p})|$, for all pixel coordinates $\vec{p}$. A binary edit mask, $M_{edit}$, is then generated by applying a threshold to this difference map, isolating the set of modified pixels over which the analysis is performed.

    \item \textbf{Local Contrast Calculation.} We define the local contrast at a pixel $\vec{p}$, denoted $C(\vec{p})$, as the standard deviation of pixel intensities within a $k \times k$ window centered at $\vec{p}$. This operation is performed for both the input and output images, yielding local contrast maps $C_{in}$ and $C_{out}$.

    \item \textbf{Gain Computation.} The final Residual Contrast Gain, $\Delta C_{res}$, is the difference between the average local contrast of the output and input images, computed exclusively over the set of edited pixels (where $M_{edit}=1$). This is formulated as:
    \begin{equation}
        \Delta C_{res} = \text{mean}_{\vec{p} | M_{edit}(\vec{p})=1} \left( C_{out}(\vec{p}) - C_{in}(\vec{p}) \right)
    \end{equation}
    A positive $\Delta C_{res}$ value indicates a net increase in detail and texture within the restored regions.
\end{enumerate}

\subsection{QwenQA: VLM-based Assessment}
Moreover, we also developed the \textbf{QwenQA} evaluation metric, to leverage the powerful Vision-Language Model (VLM) for more human-like visual assessments. Our framework is built upon the \textbf{Qwen2.5-VL-72B-Instruct} model~\citep{bai2025qwen2}. The evaluation protocol is designed for consistency and automated parsing, involving three key stages:

\begin{enumerate}
    \item \textbf{Input Standardization.} To eliminate resolution as a confounding variable, the model's prediction image is first resampled to match the exact dimensions of the original input image, ensuring a fair comparison context for the VLM.

    \item \textbf{Constrained Prompt Engineering.} The core of QwenQA lies in a meticulously engineered prompt designed to elicit a precise and quantitative response. The prompt structure includes:
    \begin{itemize}
    \item \textit{Role Assignment:} The VLM is instructed to act as a ``top-tier image quality assessment expert,'' priming it to leverage its most relevant internal knowledge.
    
    \item \textit{Task Definition:} The prompt provides clear context, defining ``Image A'' as the original with a specific artifact (e.g., 'haze', 'shadow') and ``Image B'' as the processed result.
    
    \item \textit{Objective Quantization:} The VLM's objective is narrowly focused on a single quantitative task: ``evaluate the percentage by which the '[artifact name]' is reduced in Image B compared to Image A''. This transforms a descriptive task into a quantitative one.
    
    \item \textit{Strict Output Formatting:} The prompt strictly constrains the VLM's output to a specific format: ``Score: [number]\%''. This instruction explicitly forbids any additional descriptive text, explanations, or conversational filler, which is critical for reliable automated parsing.
    \end{itemize}
    
    \item \textbf{Automated Score Parsing.} The final step is to parse the VLM's structured textual output. A regular expression is used to robustly extract the numerical percentage score from the response, yielding the final QwenQA score.
\end{enumerate}

\section{Implementation Details}

\subsection{Haze Synthesis Details}

Our primary objective in data expansion was to generate a challenging and realistic training set that surpasses the limitations of existing synthetic datasets. To achieve this, we developed a high-throughput synthesis pipeline to apply our physically-motivated atmospheric rendering model on a large scale. This section details the parameterization for various haze types, the batch processing architecture, and the datasets involved.

\noindent\textbf{Parameterization for Diverse Atmospheric Effects.}
The versatility of our rendering model allows us to simulate a wide range of atmospheric conditions by adjusting a few key physical parameters. We defined distinct configurations for haze, fog, and smoke, which were systematically varied to ensure a broad data distribution.
\begin{itemize}
    \item \textbf{Haze:} To simulate different environmental conditions, we primarily varied the \textit{airlight} color and \textit{visibility}. For instance, we used sky-tinted colors like (153, 174, 215) for typical haze, warmer tones such as (200, 180, 140) for urban pollution, and grayish colors like (210, 210, 220) for high-altitude conditions. Visibility was typically set in the range of 100m to 1000m to produce varying levels of haze density.
    
    \item \textbf{Fog:} Fog is characterized by its dense, non-absorptive particles. We simulated this by setting the single-scattering albedo $\omega_0$ to (1.0, 1.0, 1.0) and using a neutral white airlight. Fog density was controlled by varying \textit{visibility} (from 30m to 1000m) and the multiple-scattering boost exponent $\eta$ (typically between 0.5 and 1.0). To simulate low-lying or valley fog, we significantly reduced the scale height $H$ (e.g., to 30-60m) to confine the effect to the lower parts of the scene.
    
    \item \textbf{Smoke:} Unlike haze and fog, smoke is an absorptive medium. This was modeled by setting $\omega_0$ to values less than 1.0 (e.g., 0.75 to 0.85). The \textit{airlight} was configured with warm, darker colors like (180, 150, 120) or (160, 120, 90) to represent the tint of the smoke particles. The scale height $H$ was generally kept low (e.g., 40-50m) to simulate ground-level smoke plumes.
\end{itemize}

\noindent\textbf{Large-Scale Batch Synthesis Architecture.}
To apply these configurations across a massive number of images, we implemented an efficient, parallelized processing pipeline. The core rendering engine was ported to PyTorch~\citep{paszke2019pytorch} to leverage GPU acceleration. We utilized multiprocessing to create a pool of worker processes. In a multi-GPU environment, these workers were assigned to available GPUs in a round-robin fashion, enabling concurrent rendering of multiple image-configuration pairs. Each worker independently handled the data I/O, pre-processing (color space conversion, data normalization), GPU-based rendering, and post-processing of the synthesized hazy image. This architecture allowed us to generate our extensive dataset in a time-efficient manner.

\noindent\textbf{Datasets for Synthesis.}
As stated in our methodology, our goal was to enhance existing large-scale datasets by generating more challenging and realistic haze effects. We leveraged the high-quality, clean ground truth images from public benchmarks, primarily {RESIDE} \citep{li2018benchmarking} and {HAZESPACE} \citep{islam2024hazespace2m}. For each clean image in these datasets, we first estimated a monocular depth map \citep{ke2025marigold} and then applied our full suite of atmospheric rendering configurations, resulting in a significant expansion of the training data with diverse and physically plausible haze, fog, and smoke effects.

\subsection{Training Details}
Our work builds upon a pretrained DiT-based image editing model that has demonstrated strong capabilities in general inpainting tasks, such as object addition, removal, and modification. This provides a robust starting point for fine-tuning on our specialized soft-effects dataset. A key aspect of our training methodology is a hierarchical data sampling strategy designed to balance contributions from numerous datasets across multiple tasks. Our data pipeline first groups datasets by their primary task (e.g., shadow removal, dehazing, reflection removal, etc.). During each training step, a task is uniformly sampled, and then a specific dataset within that task group is selected based on a predefined sampling weight. This weighting ratio is configured for each dataset, allowing us to strategically oversample smaller, high-quality real-world datasets to learn the knowledge without domain gaps, while still benefiting from the diversity of larger-scale synthetic data sources to prevent overfitting and enhance the generalization ability. This ensures the model receives a balanced and comprehensive exposure to all types of soft effects.

For the fine-tuning process, our model operates within the DDPM~\citep{ho2020denoising} framework, which is adapted to use continuous timesteps for increased flexibility. Notably, we employ $\upsilon$-parameterization instead of the standard $\epsilon$-parameterization to improve training stability and sample quality. Our training objective is to predict the noise added to the clean image's latent representation at a given timestep. The loss function is the mean squared error (MSE) between the predicted noise and the ground truth noise, with a timestep-dependent weighting scheme applied to balance the contribution of different noise levels throughout the training. We train the model for 10k steps at a resolution of 1024x1024. We employ the AdamW optimizer with a learning rate of $1.2 \times 10^{-5}$, governed by a linear warmup of 2000 steps followed by a cosine decay schedule. Our UniSER is trained on all of the data mentioned above with $8$ NVIDIA A100 80G for 10k iterations.

\subsection{Evaluation Details for Baselines}
When evaluating the generalist baselines, we provided detailed and specific text prompts to ensure they could achieve their optimal performance. These prompts explicitly described the effect to be removed and the relevant scene context, for instance: \textit{''remove the atmosphere haze completely in this image''} or \textit{''remove the shadow casted by the giraffe on the grass''}. Furthermore, to account for the stochastic nature of generative models, if a model performed poorly or failed to remove the effect on a particular sample, we conducted multiple attempts to ensures we are not using ambiguous or vague text prompts. This is a fair evaluation and mitigates biases arising from individual random outcomes. In contrast, our UniSER has minimal dependency on text prompts. In our framework, the text serves merely as a high-level task indicator (e.g., \textit{''remove haze''}) without requiring a detailed description of the scene's content. Consequently, our approach achieves stable and robust results without the need for iterative prompt engineering.

\section{More Visual Results and Quality Analysis}

\subsection{More Visual Results}
We provide more visual results in Fig.~\ref{fig:gallery_remove1}, Fig.~\ref{fig:gallery_remove2} and Fig.~\ref{fig:gallery_add}, by randomly pick in-the-wild photos degraded by soft effects, our UniSER shows perfect robustness on thoroughly removing the. Besides, UniSER is also capable of generating or enhancing multiple effects aesthetically.

\subsection{Contrast Analysis}
To further investigate how UniSER improves image quality, we visualize the local contrast maps of images before and after editing, as shown in Figure~\ref{fig:additional}. A significant enhancement in contrast is observed within the regions originally degraded by soft effects. This indicates that our method not only removes the obstructive artifacts but also successfully restores and enhances the underlying image details and textures that were suppressed by the effects, leading to a clearer and more vivid output.

\end{document}